\documentclass[letterpaper, 10 pt, conference]{ieeeconf}  

\IEEEoverridecommandlockouts                              

\usepackage{graphics} 
\usepackage{epsfig} 
\usepackage{amsmath} 
\usepackage{amssymb}  

\usepackage{xcolor}

\usepackage{ulem} 
\usepackage{kotex} 

\title{\LARGE \bf
Haptic-Based Bilateral Teleoperation of Aerial Manipulator for Extracting Wedged Object with Compensation of Human Reaction Time
}

\author{Jeonghyun Byun$^{1}$, Dohyun Eom$^{1}$, and H. Jin Kim$^{1}$
\thanks{This research was supported by Unmanned Vehicles Core Technology Research and Development Program through the National Research  Foundation of Korea(NRF) and Unmanned Vehicle Advanced Research Center(UVARC) funded by the Ministry of Science and ICT, the Republic of Korea(NRF-2020M3C1C1A010864).}
\thanks{$^{1}$ The authors are with the Department of Aerospace Engineering,  Automation and System Research Institute(ASRI) and Institute of Advanced Aerospace Technology(IAAT), Seoul National University,  Seoul, South Korea.
        {\tt\small \{quswjdgus97, djaehgus, hjinkim\}@snu.ac.kr}}}

\begin{document}

\maketitle
\thispagestyle{empty}
\pagestyle{empty}

\begin{abstract}

Bilateral teleoperation of an aerial manipulator facilitates the execution of industrial missions thanks to the combination of the aerial platform's maneuverability and the ability to conduct complex tasks with human supervision. 
Heretofore, research on such operations has focused on flying without any physical interaction or exerting a pushing force on a contact surface that does not involve abrupt changes in the interaction force. 
In this paper, we propose a human reaction time compensating haptic-based bilateral teleoperation strategy for an aerial manipulator extracting a wedged object from a static structure (i.e., plug-pulling), which incurs an abrupt decrease in the interaction force and causes additional difficulty for an aerial platform.
A haptic device composed of a 4-degree-of-freedom robotic arm and a gripper is made for the teleoperation of aerial wedged object-extracting tasks, and a haptic-based teleoperation method to execute the aerial manipulator by the haptic device is introduced.
We detect the extraction of the object by the estimation of the external force exerted on the aerial manipulator and generate reference trajectories for both the aerial manipulator and the haptic device after the extraction.
As an example of the extraction of a wedged object, we conduct comparative plug-pulling experiments with a quadrotor-based aerial manipulator. 
The results validate that the proposed bilateral teleoperation method reduces the overshoot in the aerial manipulator's position and ensures fast recovery to its initial position after extracting the wedged object.

\end{abstract}

\section{INTRODUCTION}

Unmanned aerial manipulators (UAMs) have gained increasing attention owing to their maneuverability and versatility. Thanks to the maneuverability, they can be utilized in places that are dangerous for human operators or inappropriate for ground-based robots including walls or windows located in tall buildings, wind turbines, decommissioned reactors, or power transmission towers. Also, thanks to their versatility, there exist several studies on a UAM conducting physically interacting tasks such as drawer opening \cite{kim2015operating}, peg-in-hole operation \cite{car2018impedance}, non-destructive inspection \cite{trujillo2019novel}, door opening \cite{lee2020aerial} and window-cleaning \cite{sun2021switchable}. 

Even though there exist works on fully autonomous control of UAMs \cite{kim2017robust, liang2021low, lee2022rise}, these controllers might face safety issues while operating in a priorly unknown environment or require complicated algorithms with high computational power to perform complex tasks. To resolve these problems, the ``human-in-the-loop" control of a UAM is adopted to utilize humans' decision-making ability while conducting complex tasks. Also, sensory information gathered by the UAM can be transmitted to the human operator to aid the decision-making process. Hence, research on the control of a UAM using the mutual exchange of information between the human operator and the aerial vehicle, or \textit{bilateral teleoperation} \cite{mersha2013bilateral}, is meaningful. 

\begin{figure}[!t] 
    \centering
    \includegraphics[width=0.46\textwidth]{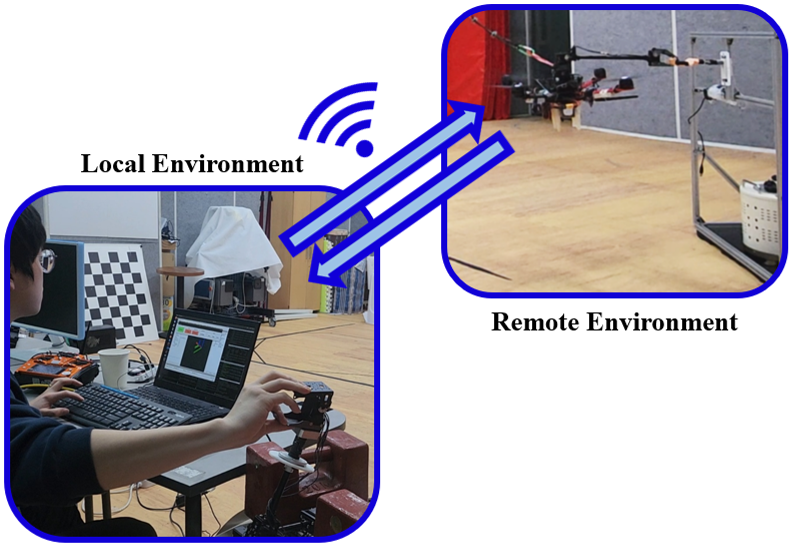}
    \caption{Illustration of the haptic-based bilateral teleoperation of a UAM (remote environment) extracting a wedged object from a static structure with a haptic device (local environment).}
    \label{fig: thumbnail}
\end{figure}


However, there are a few situations in which a human operator cannot quickly pilot the UAM due to the limitation of the humans' reaction time. Those situations include the tasks involving abrupt decrease in the interaction force such as extracting a wedged object from a static structure \cite{byun2023hybrid}. When those situations occur, the instantaneous fully autonomous control of the UAM and the haptic device after the abrupt force change can prevent an excessive overshoot in the UAM's position or system destabilization induced by the operator's long reaction time.

This paper proposes a haptic-based bilateral teleoperation of an aerial manipulator for extracting a wedged object from a static structure with the compensation of human reaction time, as shown in Fig. \ref{fig: thumbnail}. 
To this end, we fabricate a haptic device configured with a 4-degree-of-freedom (DOF) and a gripper for grasping and pulling a wedged object using a UAM.
Then, we divide the aerial object-extracting task into two flight phases including nominal and recovery phases where the recovery flight phase refers to a short time interval after object extraction while the nominal flight phase means the remaining interval.
Also, the reference trajectory generation methods for the haptic device and the UAM during two flight phases are designed and the detection algorithm for switching from the nominal to the recovery flight phase is developed. 
As an example of the wedged object-extracting tasks, we conduct two plug-pulling experiments using a quadrotor-based aerial manipulator with and without the proposed bilateral teleoperation method, and we show that the proposed method enhances the behavior of the UAM after the object extraction in the perspective of reducing the overshoot in the UAM’s position and the time duration for the UAM to return to its original position.

\subsection{Related Works}

Several studies on the bilateral teleoperation of aerial vehicles for missions with no physical interaction have been conducted. 
\cite{stramigioli2010novel} and \cite{macchini2020hand} present the teleoperation method using the haptic feedback reflecting obstacles nearby.
In \cite{masone2018shared}, the human operator indirectly guides a micro aerial vehicle by adjusting parameters related to its path generation with haptic feedback including information about tracking performance or obstacle presence.
In \cite{coelho2020whole}, the authors introduce the teleoperation approach which transmits a wall-contact feeling to a human operator when the base approaches an unreachable position.

There also exist a few works on the bilateral teleoperation of UAMs involving contact with the surrounding environment.
In \cite{gioioso2015force}, a method for the teleoperation of a UAM is introduced to enable a smooth transition between the free-flight and force-control phases and the regulation of the contact force exerted on the UAM's end-effector. 
Hereby, the haptic feedback to a human operator includes the contact force that UAM receives.
In \cite{lee2020visual}, the human operator leverages haptic feedback and 3D visual feedback provided by virtual reality for the teleoperation of a UAM conducting several aerial manipulation tasks including grasping, pick-and-place, force exertion, and peg-in-hole insertion.
\cite{kim2020human} introduces a haptic device for the teleoperation of aerial drilling, and in \cite{allenspach2022towards}, a method for the teleoperation of a 6-DOF fully actuated UAM that physically interacts with the surrounding environment is designed and tested. 
However, those works focus on the interaction tasks only involving the pushing motion where the effect of the interaction force change can be mitigated by slowing down before the physical contact.

\subsection{Contributions}

To the best of the authors' knowledge, it is the first attempt to conduct the haptic-based bilateral teleoperation of a UAM extracting a wedged object from a static structure. 
Also, we produce the haptic device configured with a 4-DOF robotic arm and a gripper for the emulation of the movement of the UAM grabbing and extracting an object wedged in a static structure. 
Moreover, an admittance controller that makes the haptic device compliant with the external torque exerted by a human's hand is designed without a force/torque sensor. 
Furthermore, to avoid destabilization or an excessive overshoot in the position of the UAM after extracting the wedged object, we design an algorithm for the detection of the object extraction and the trajectory generation method for the UAM and the haptic device after the detection. 
Lastly, for the validation of the proposed method, we conduct an actual plug-pulling experiment with a quadrotor-based aerial manipulator with our haptic device.

\subsection{Outline}

In Section II, we formulate two dynamic equations of the haptic device and the UAM, and describe the bilateral teleoperation method during the nominal flight phase in Section III. 
Section IV presents the conditions for detecting the extraction of a wedged object and the trajectory generation methods during the recovery flight.
In Section VI, the proposed teleoperation strategy is validated experimentally.

\textbf{Notations:} $\boldsymbol{0_{i\times j}}$, $\boldsymbol{I_{i}}$ and $\boldsymbol{e_3}$ represent the $i\times j$ zero matrix, $i \times i$ identity matrix and $[0 \ 0\ 1]^{\top}$, respectively. For scalars $a$, $b$ and $c$, we let $[a;b;c]$ denote $[a \ b \ c]^{\top}$. Also, $\|\boldsymbol{\alpha}\|$ and $\alpha_i$ mean the Euclidean norm and the $i$-th element of a vector $\boldsymbol{\alpha}$, respectively. Moreover, $c_{*}$ and $s_{*}$ are the abbreviations of $\cos{*}$ and $\sin{*}$, respectively.

\section{MODELING}

For the bilateral teleoperation of a UAM extracting a wedged object, we manipulate the UAM consisting of a multirotor and a 2-DOF robotic arm with a gripper using the haptic device configured with a 4-DOF robotic arm and a 1-DOF gripper.
In this section, the dynamic models of the haptic device and the UAM are derived.

\subsection{Haptic Device}

\begin{figure}[!t] 
    \centering
    \includegraphics[width=0.48\textwidth]{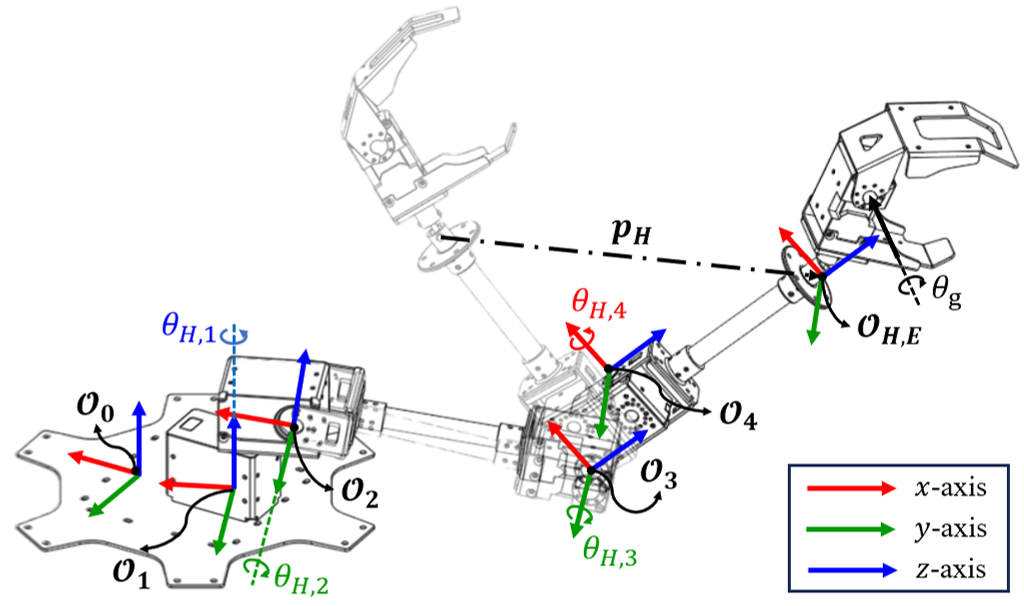}
    \caption{Illustration of the robotic arm that is used for our haptic device. Key coordinate frames are also depicted.}
    \label{fig: illustration of haptic device}
\end{figure}

In Fig. \ref{fig: illustration of haptic device}, $\boldsymbol{\mathcal{O}_{i}}$-$\boldsymbol{xyz}$ ($i = 0,\cdots,4$) and $\boldsymbol{\mathcal{O}_{H,E}}$-$\boldsymbol{xyz}$ are the coordinate frames of the $i^{\textrm{th}}$ joint and the robotic arm's end-effector, respectively. 
Also, $\boldsymbol{p_H}$ means the displacement from the initial position of $\boldsymbol{O_{H,E}}$ to its current position with respect to $\boldsymbol{\mathcal{O}_0}$-$\boldsymbol{xyz}$. 
The generalized coordinate of the robotic arm is $\boldsymbol{\theta_H} \triangleq [\theta_{H,1};\theta_{H,2};\theta_{H,3};\theta_{H,4}]$, and the joint angle of the gripper is defined as $\theta_g$.

Since a ball joint is attached between the robotic arm and the gripper, the human hand freely 
rotates. Then, we can derive dynamic models of the robotic arm and the gripper, respectively. 
For the derivation of the robotic arm's dynamic model, we assume that the gripper is considered as a point mass attached on $\boldsymbol{O_{H,E}}$.
Hence, $\boldsymbol{\ddot{\theta}_H}$ satisfies the following equation:
\begin{multline}
    \boldsymbol{M_H}(\boldsymbol{\theta_H})\boldsymbol{\ddot{\theta}_H} + \boldsymbol{C_H}(\boldsymbol{\theta_H},\boldsymbol{\dot{\theta}_H})\boldsymbol{\dot{\theta}_H} + \boldsymbol{G_H}(\boldsymbol{\theta_H}) \\= \boldsymbol{\tau_H} + \boldsymbol{\tau_{H,E}}    
\end{multline}
where $\boldsymbol{M}(\boldsymbol{\theta_H}) \in {\mathbb{R}}^{4\times4}$, $\boldsymbol{C_H}(\boldsymbol{\theta_H},\boldsymbol{\dot{\theta}_H}) \in {\mathbb{R}}^{4\times4}$ and $\boldsymbol{G_H}(\boldsymbol{\theta_H}) \in {\mathbb{R}}^{4}$ are the mass, Coriolis and gravity matrices of the robotic arm, respectively, and we let $\boldsymbol{\tau_H} \in {\mathbb{R}}^4$ and $\boldsymbol{\tau_{H,E}} \in {\mathbb{R}}^4$ denote the joint torque vector and the external torque applied by the human operator, respectively.

Meanwhile, the model of the gripper is:
\begin{equation} 
    J_g\ddot{\theta}_g = \tau_g
\end{equation}
where $J_g$ and $\tau_g$ are the moment of inertia of the gripper's servo motor with respect to the rotation axis and the corresponding joint torque, respectively.

\subsection{UAM}

\begin{figure}[!t] 
    \centering
    \includegraphics[width=0.44\textwidth]{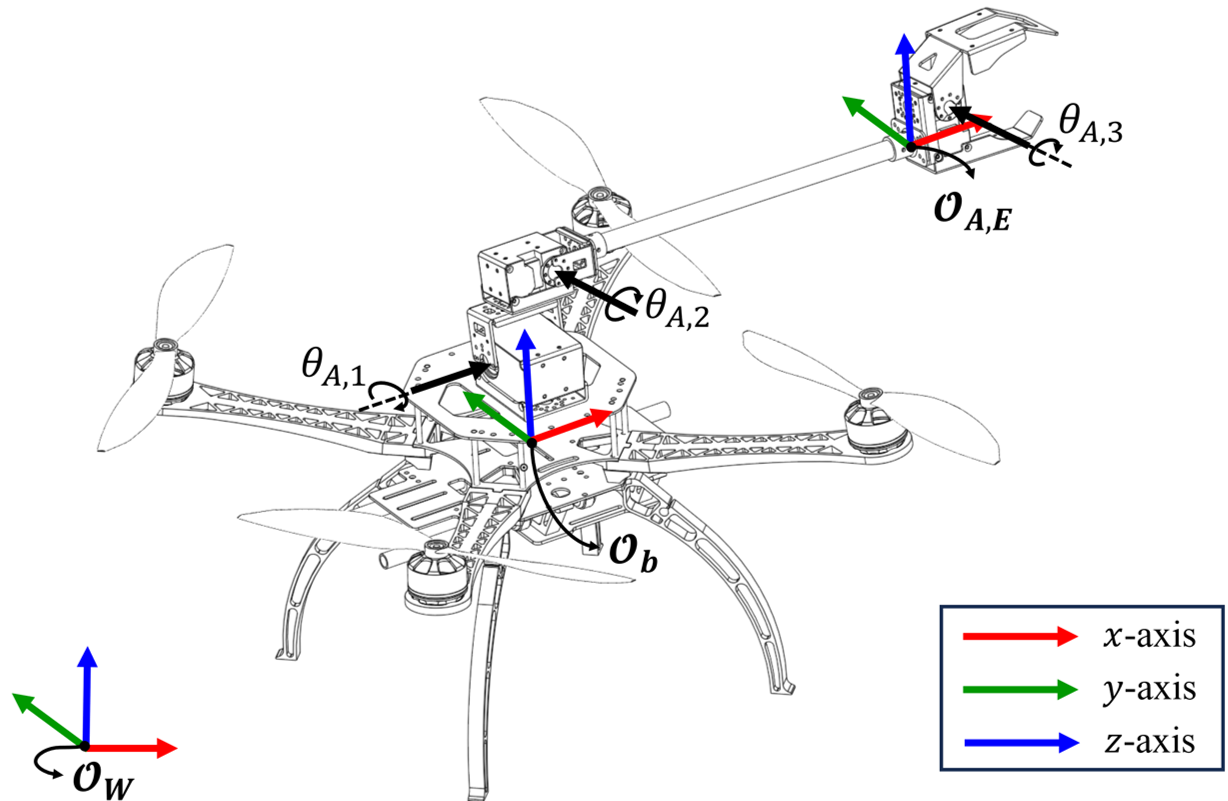}
    \caption{Illustration of the UAM with the key coordinate frames.}
    \label{fig: illustration of UAM}
\end{figure}

The key coordinate frames of the UAM are shown in Fig. \ref{fig: illustration of UAM}. 
$\boldsymbol{\mathcal{O}_W}$-$\boldsymbol{xyz}$, $\boldsymbol{\mathcal{O}_b}$-$\boldsymbol{xyz}$ and $\boldsymbol{\mathcal{O}_{A,E}}$-$\boldsymbol{xyz}$ represent the coordinate frames of the world, multirotor, and the end-effector of the UAM, respectively. 
In this paper, the translational movement of the vehicle is described by the position of the UAM's center of mass, $\boldsymbol{p_c}$, and its rotational movement is expressed with the Euler angles of the multirotor, $\boldsymbol{\phi} \in {\mathbb{R}}^3$. 
Also, the joint angles of the attached robotic arm are defined as $\boldsymbol{\theta_a} \in {\mathbb{R}}^3$. 

According to \cite{yang2014dynamics}, the dynamic model with respect to $\boldsymbol{p_c}$ is formulated as follows:
\begin{equation} \label{eqn: translational dynamic model of UAM}
    \begin{split}
        m_t\boldsymbol{\ddot{p}_c} =& -m_tg\boldsymbol{e_3} + \boldsymbol{u_c} + \boldsymbol{R^{\top}_b}\boldsymbol{f_{A,E}}
    \end{split}
\end{equation}
where $m_t$ and $g$ are the total mass of the UAM and the gravitational acceleration, respectively, $\boldsymbol{R_b} \in SO(3)$ means the rotational matrix from $\boldsymbol{O_b}$-$\boldsymbol{xyz}$ to $\boldsymbol{O_W}$-$\boldsymbol{xyz}$, and
\begin{equation*}
    \begin{split}
        \boldsymbol{u_c} =& F\underset{\triangleq \boldsymbol{\Psi}}{\underline{\begin{bmatrix}
            c\phi_3 & s\phi_3 & 0 \\
            s\phi_3 & -c\phi_3 & 0 \\
            0 & 0 & 1
        \end{bmatrix}}}\begin{bmatrix}
            c\phi_1 s\phi_2 \\ 
            s\phi_1 \\
            c\phi_1 c\phi_2
        \end{bmatrix}
    \end{split}
\end{equation*}
with $F$ representing the total thrust of the UAM. 
Also, $\boldsymbol{f_{A,E}} \in {\mathbb{R}}^3$ represents the force exerted on the end-effector expressed in $\boldsymbol{O_b}$-$\boldsymbol{xyz}$. 

Meanwhile, with the assumption that the attitude of the multirotor and the joint angles of the robotic arm are well controlled by their built-in controllers (e.g., PX4, Dynamixel), the control inputs of the multirotor's attitude and the joint angles become the desired angular velocity of the multirotor, $\boldsymbol{\omega_{b,d}}$, and the desired joint angles, $\boldsymbol{\theta_{A,d}} \in {\mathbb{R}}^3$, respectively.
Hence, the dynamic equations with respect to $\boldsymbol{\phi}$ and $\boldsymbol{\theta}_A$ are expressed as follows:
\begin{equation}
    \begin{split}
        \boldsymbol{\dot{\phi}} =& \boldsymbol{Q}^{-1}(\boldsymbol{\phi})\boldsymbol{\omega_{b,d}}, \quad \boldsymbol{{\theta}_A} = \boldsymbol{{\theta}_{A,d}}
    \end{split}
\end{equation}
where $\boldsymbol{Q}(\boldsymbol{\phi}) \triangleq \begin{bmatrix}
            1 & 0 & -s\phi_2 \\
            0 & c\phi_1 & s\phi_1 c\phi_2 \\
            0 & -s\phi_1 & c\phi_1 c\phi_2
        \end{bmatrix}$.

\section{BILATERAL TELEOPERATION DURING NOMINAL FLIGHT}

\begin{figure*}[!t] 
    \centering
    \includegraphics[width=0.98\textwidth]{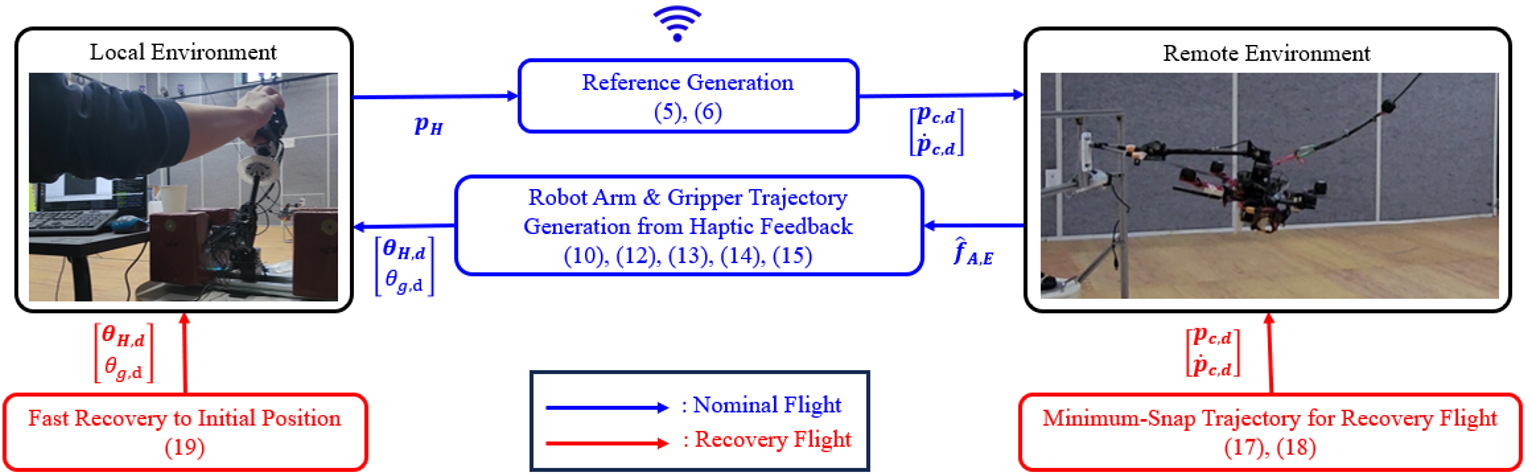}
    \caption{Diagram including the teleoperation method during nominal flight (blue lines) and reference trajectory generation methods for the haptic device and the UAM during recovery flight (red lines).}
    \label{fig: entire control flow}
\end{figure*}

The \textit{nominal flight phase} means that there does not occur sudden change in $\boldsymbol{f_{A,E}}$, i.e., it does not involve the extraction of a wedged object from a static structure. 
In this section, the bilateral teleoperation strategy during the nominal flight phase based on \cite{allenspach2022towards} is presented. 
The overall flow map of this phase is described in Fig. \ref{fig: entire control flow} with the blue lines.



\subsection{UAM}

\subsubsection{Reference Trajectory Generation}

From the displacement of the haptic device's tooltip, $\boldsymbol{p_H}$, the desired velocity of the UAM's COM with respect to $\boldsymbol{O_b}$-$\boldsymbol{xyz}$, $\boldsymbol{\dot{p}^b_{c,d}}$, is designed as follows:
\begin{equation} \label{eqn: desired velocity of the UAM's c.o.m. w.r.t the body frame}
    \begin{split}
        \boldsymbol{\dot{p}^b_{c,d}} = \begin{bmatrix}
            v_{\textrm{max},x}\tanh{(\tfrac{p_{H,1}}{p_{H,\textrm{max},x}})} \\
            v_{\textrm{max},y}\tanh{(\tfrac{p_{H,2}}{p_{H,\textrm{max},y}})} \\
            v_{\textrm{max},z}\tanh{(\tfrac{p_{H,3}}{p_{H,\textrm{max},z}})}
        \end{bmatrix}.
    \end{split}
\end{equation}
where $v_{\textrm{max},x}$, $v_{\textrm{max},y}$ and $v_{\textrm{max},z}$ represent the UAM's maximum speeds of $x$, $y$ and $z$-axes of the body frame, respectively, and we let $p_{H,\textrm{max},x}$, $p_{H,\textrm{max},y}$ and $p_{H,\textrm{max},z}$ denote the maximum absolute values of each element of $\boldsymbol{p_H}$.
Then from the above equation, the desired velocity and position of $\boldsymbol{p_c}$ are obtained as follows:
\begin{equation}
    \begin{split}
        \boldsymbol{\dot{p}_{c,d}} =& \boldsymbol{R_z}(\phi_3)\boldsymbol{\dot{p}^b_{c,d}} \\
        \boldsymbol{p_{c,d}} =& \boldsymbol{p_{c,0}} + \int^{t}_{t_0} \boldsymbol{\dot{p}_{c,d}}(\tau) d\tau
    \end{split}
\end{equation}
where $\boldsymbol{p_{c,0}}$ is the initial value of $\boldsymbol{p_c}$.

While the desired yaw angle of the multirotor, $\phi_{d,3}$, is set from the keyboard of the ground control station by the human operator, the desired joint angles are set as follows:
\begin{equation} \label{eqn: desired joint angles}
    \begin{split}
        \boldsymbol{\theta_{A,d}} = [-\phi_1;-\phi_2;\theta_g].
    \end{split}
\end{equation}
This equation makes the attitude of the end-effector parallel to the ground and commands the UAM's gripper to follow the signal from the haptic device.


\subsubsection{Control}

For the regulation of $\boldsymbol{p_c}$, we adopt the disturbance-observer (DOB)-based controller introduced in \cite{ha2018disturbance} to generate the desired values of the total thrust $F$, the roll angle $\phi_1$ and the pitch angle $\phi_2$ from the error between $\boldsymbol{p_c}$ and $\boldsymbol{p_{c,d}}$. 
Then, based on $\boldsymbol{\phi_d} = [\phi_{d,1};\phi_{d,2};\phi_{d,3}]$, we calculate $\boldsymbol{\omega^b_{b,d}}$ for the control of the multirotor's attitude. 
Lastly, we simply transmit $\boldsymbol{\theta_{A,d}}$ in (\ref{eqn: desired joint angles}) to the built-in controller of the servo motors. 

The desired value of $\boldsymbol{u_c}$ in (\ref{eqn: translational dynamic model of UAM}), $\boldsymbol{u_{c,d}}$, is calculated as follows:
\begin{multline} \label{eqn: position control of UAM}
    \boldsymbol{u_{c,d}} = \hat{m}_t(\hat{g}\boldsymbol{e_3} \\ - \boldsymbol{K_D}(\boldsymbol{\dot{p}_c} - \boldsymbol{\dot{p}_{c,d}}) - \boldsymbol{K_P}(\boldsymbol{p_c} - \boldsymbol{p_{c,d}})) - \boldsymbol{R_b}\boldsymbol{\hat{f}_{A,E}}
\end{multline}
where $\hat{m}_t$ and $\hat{g}$ are the nominal values of $m_t$ and $g$, respectively, and $\boldsymbol{\hat{f}_{A,E}}$ is the estimated value of $\boldsymbol{f_{A,E}}$ which will be presented later. 
Then, $F$ and the desired roll and pitch angles, $\phi_{d,1}$ and $\phi_{d,2}$, are extracted from $\boldsymbol{u}_{c,d}$ as follows:
\begin{equation}
    \begin{split}
        F =& \tfrac{1}{c\phi_1 c\phi_2}\boldsymbol{e_3^{\top}}\boldsymbol{\Psi}\boldsymbol{u_{c,d}}, \ \phi_{d,1} = \sin^{-1}{(\tfrac{1}{F}\boldsymbol{e_2^{\top}}\boldsymbol{\Psi}\boldsymbol{u_{c,d}})}, \\
        \phi_{d,2} =& \sin^{-1}{(\tfrac{1}{F c\phi_1}\boldsymbol{e_1^{\top}}\boldsymbol{\Psi}\boldsymbol{u_{c,d}})}
    \end{split}
\end{equation}

Meanwhile, $\boldsymbol{\hat{f}_{A,E}}$ is calculated based on $\boldsymbol{u_c}$ as follows:
\begin{equation} \label{eqn: disturbance observer for the locked space}
    \begin{split}
        \boldsymbol{\dot{\zeta}_c} =& -\tfrac{1}{\nu_c}\boldsymbol{\Gamma_{c,\zeta}}(\boldsymbol{\zeta_c} -\boldsymbol{\dot{p}_{c}}), \qquad \ \ \boldsymbol{\zeta_{c}}(t_0) = \boldsymbol{p_{c,0}} \\
        \boldsymbol{\dot{\chi}_c} =& -\tfrac{1}{\nu_c}\boldsymbol{\Gamma_{c,\chi}}(\boldsymbol{\chi_c} - \tfrac{1}{\hat{m}_t}\boldsymbol{u_c}), \quad \boldsymbol{\chi_{c}}(t_0) = g\boldsymbol{e_3} \\
        \boldsymbol{\hat{f}_{A,E}} =& -\tfrac{\hat{m}_t}{\nu_c}\boldsymbol{\Gamma_{c,\zeta}}(\boldsymbol{\zeta_c} - \boldsymbol{\dot{p}_c}) + \hat{m}_t\hat{g}\boldsymbol{e_3} - \hat{m}_t\boldsymbol{\chi_c}
    \end{split}
\end{equation}
where $\boldsymbol{\zeta_c} \in {\mathbb{R}}^{3}$ and $\boldsymbol{\chi_c} \in {\mathbb{R}}^{3}$ represents the estimation of $\boldsymbol{\dot{p}_c}$ and the filtered value of $\tfrac{1}{\hat{m}_t}\boldsymbol{u_c}$, respectively, and $\boldsymbol{\Gamma_{c,\zeta}}$, $\boldsymbol{\Gamma_{c,\chi}} \in {\mathbb{R}}^{3\times3}$ and $\nu_c < 1$ are the user-defined positive definite matrices and positive scalar value smaller than 1. 
Note that $\boldsymbol{\hat{f}_{A,E}}$ is calculated based on $\boldsymbol{u_c}$, not $\boldsymbol{u_{c,d}}$, to exclude the effects of the difference between $\boldsymbol{u_c}$ and $\boldsymbol{u_{c,d}}$.

Meanwhile, $\boldsymbol{\omega_{b,d}}$ is obtained as follows:
\begin{equation} \label{eqn: attitude control of multirotor}
    \begin{split}
        \boldsymbol{\omega_{b,d}} = \boldsymbol{K_R}(\boldsymbol{\phi_d} - \boldsymbol{\phi})
    \end{split}
\end{equation}
where $\boldsymbol{K_R} \in {\mathbb{R}}^{3 \times 3}$ is the user-defined positive definite matrix, and we directly input $\boldsymbol{\theta_{A,d}}$ generated in (\ref{eqn: desired joint angles}) into the built-in controller to control the joint angle control.

\subsection{Haptic Device: Robotic Arm}

Let the positive definite matrices $\boldsymbol{M_{H,d}} \in {\mathbb{R}}^{4\times4}$ and $\boldsymbol{D_{H,d}} \in {\mathbb{R}}^{4 \times 4}$ denote the desired inertia and damping matrices of the haptic device, respectively, then the admittance control law for $\boldsymbol{\theta_H}$ is designed as follows:
\begin{multline} \label{eqn: control of robotic arm of haptic device during nominal flight}
    \boldsymbol{M_{H,d}}\boldsymbol{\ddot{\theta}_{H,d}} + \boldsymbol{D_{H,d}}\boldsymbol{\dot{\theta}_{H,d}} \\= \boldsymbol{\hat{\tau}_{H,E}} + \boldsymbol{\tau_{fb}} + \boldsymbol{J_{H,E}}(\boldsymbol{\theta_H})\boldsymbol{\hat{f}_{A,E}} 
\end{multline}
where we let $\boldsymbol{\hat{\tau}_{H,E}} \in {\mathbb{R}}^4$ and $\boldsymbol{\tau_{fb}} \in {\mathbb{R}}^4$ denote the estimated value of $\boldsymbol{\tau_{H,E}}$ and recentering torque, respectively.

\subsubsection{Estimation of External Torque Applied to Haptic Device}

The external torque from the human operator is estimated by the momentum-based estimator introduced in \cite{tomic2017external} as follows:
\begin{multline}
    \boldsymbol{\hat{\tau}_{H,E}} = \boldsymbol{K_{I,H}}\Bigg[\boldsymbol{M}(\boldsymbol{\theta_H})\boldsymbol{\dot{\theta}_H} - \int^t_{t_0}(\boldsymbol{\tau_H} - \boldsymbol{C}(\boldsymbol{\theta_H},\boldsymbol{\dot{\theta}_H})\boldsymbol{\dot{\theta}_H} \\ -\boldsymbol{G}(\boldsymbol{\theta_H}) + \boldsymbol{\hat{\tau}_{H,E}})d\tau - \boldsymbol{M}(\boldsymbol{\theta_{H,0}})\boldsymbol{\dot{\theta}_{H,0}}\Bigg], \ \boldsymbol{\hat{\tau}_{H,E,0}} = \boldsymbol{0_{4\times1}}    
\end{multline}
where $\boldsymbol{K_{I,H}} \in {\mathbb{R}}^{4 \times 4}$ is the user-defined positive definite matrix and we let $\boldsymbol{\theta_{H,0}}$ and $\boldsymbol{\hat{\tau}_{H,E,0}}$ denote the initial values of $\boldsymbol{\theta_H}$ and $\boldsymbol{\hat{\tau}_{H,E}}$, respectively.

\subsubsection{Recentering Torque}

To drive the haptic device to its initial state when $\boldsymbol{\tau_{H,E}}$ is $\boldsymbol{0_{3\time1}}$, the recentering torque, $\boldsymbol{\tau_{fb}} \in {\mathbb{R}}^4$, is designed as follows:
\begin{equation}
    \begin{split}
        \boldsymbol{\tau_{fb}} =& -\boldsymbol{K_{fb}}(\boldsymbol{\theta_H} - \boldsymbol{\theta_{H,0}})
    \end{split}
\end{equation}
where $\boldsymbol{K_{fb}} \in {\mathbb{R}}^{4\times4}$ represents a user-defined positive definite matrix

\subsection{Haptic Device: Gripper}

To make the gripper of the haptic device compliant with the grasping motion of the human hand, the desired angle of the gripper, $\theta_{g,d}$, is updated as follows:
\begin{equation}
    \begin{split}
        \ddot{\theta}_{g,d} = -k_{d,g}\dot{\theta}_{g,d} - k_{fb,g}(\theta_g - \theta_{g}(t_0)) + k_{\tau}\tau_g  
    \end{split}
\end{equation}
with the positive parameters $k_{d,g}$, $k_{fb,g}$ and $k_{\tau}$

\section{RECOVERY FLIGHT}

When the extraction of the wedged object is detected, an excessive overshoot in the magnitude of $\boldsymbol{p_c}$ might occur since a human operator cannot quickly react to the extraction of a wedged object.
Hence, the \textit{recovery flight}, which means the fully autonomous maneuver of the UAM returning to its state at the moment of extraction, is necessary to avoid the destabilization of the UAM or the large magnitude of position overshoot.

In this section, a method for detecting a wedged object extraction is introduced.
Then, we present the methods of reference trajectory generation for the UAM and the haptic device during the recovery flight phase. 
During this phase, the UAM and the haptic device do not mutually exchange information and they are controlled fully autonomously. 
The detailed description is shown in Fig. \ref{fig: entire control flow} with the red lines.

Note that $T_e$ and $t_e$ represent the user-defined time duration of the recovery flight and the time instant of the wedged object extraction, respectively.

\subsection{Detection of Wedged Object Extraction}

By using the linear Kalman filter, we calculate $\boldsymbol{\dot{\hat{f}}_{A,E}}$. 
Then, we can determine whether the wedged object is extracted or not by examining the decrease speed of $\|\boldsymbol{\dot{\hat{f}}_{A,E}}\|$. 
From this concept, the switching from the nominal to recovery flight is detected by a condition shown below:
\begin{equation}
    \begin{split}
        \hat{\dot{f}}_{A,E,thres} \leq \|\boldsymbol{\dot{\hat{f}}_{A,E}}\| 
    \end{split}
\end{equation}
where $\hat{\dot{f}}_{A,E,thres}$ represents the minimum threshold.

\subsection{Reference Trajectory Generation for UAM}

According to \cite{liu2021toward}, the motor commands and the acceleration of the Euler angles are known to be proportional to the snap of the position trajectory. 
Therefore, we design a minimum-snap trajectory of $\boldsymbol{p_c}$ for the recovery flight as follows:
\begin{equation} \label{eqn: minimum-snap trajectory}
    \begin{split}
        \boldsymbol{p_{c,d}} = \Sigma^{7}_{i=0}\boldsymbol{c_i}t^i, \quad \boldsymbol{\dot{p}_{c,d}} = \Sigma^{7}_{i=1}i\boldsymbol{c_i}t^{i-1}
    \end{split}
\end{equation}
where the coefficient $\boldsymbol{c_i}$ is calculated by solving the optimization problem below:
\begin{equation}
    \begin{split}
        \underset{\boldsymbol{c_0},\cdots,\boldsymbol{c_7}}{\textrm{min}} & \int^{t_e + T_e}_{t_e} \|\boldsymbol{\ddddot{p}_{c,d}}\|^2 d\tau \\
        \textrm{s.t.} \ & \boldsymbol{p_{c,d}}(t_e) = \boldsymbol{p_e}, \quad \boldsymbol{p_{c,d}}(t_e + T_e) = \boldsymbol{p_e} \\
        & \boldsymbol{\dot{p}_{c,d}}(t_e) = \boldsymbol{\dot{p}_e}, \quad \boldsymbol{\dot{p}_{c,d}}(t_e + T_e) = \boldsymbol{0_{3\times1}} \\
        & \boldsymbol{\ddot{p}_{c,d}}(t_e + T_e) = \boldsymbol{\dddot{p}_{c,d}}(t_e + T_e) = \boldsymbol{0_{3\times1}}
    \end{split}
\end{equation}
where $\boldsymbol{p_e}$ and $\boldsymbol{\dot{p}_e}$ represent the values of $\boldsymbol{p_c}$ and $\boldsymbol{\dot{p}_c}$ at $t = t_e$.

Meanwhile, $\phi_{d,3}$, $\theta_{A,d,1}$ and $\theta_{A,d,2}$ are the same as the value during the nominal flight. 


For the control of the UAM, we use the same control laws introduced in (\ref{eqn: position control of UAM}) - (\ref{eqn: attitude control of multirotor}).

\subsection{Reference Trajectory Generation for Haptic Device}

If there exists an error between $\boldsymbol{\theta_{H,d}}$ and $\boldsymbol{\theta_H}$ at the moment of re-switching from the recovery to nominal phases, the UAM will face a sudden movement. 
Hence, the robotic arm of the haptic device is commanded to quickly recover its initial pose during the recovery flight as follows:
\begin{equation}
    \begin{split}
        \boldsymbol{\dot{\theta}_{H,d}} = -\boldsymbol{K_{recovery}}(\boldsymbol{\theta_H} - \boldsymbol{\theta_{H,0}})
    \end{split}
\end{equation}
with the user defined positive definite matrix $\boldsymbol{K_{recovery}} \in {\mathbb{R}}^{4\times4}$. 

To hold onto the extracted object during the recovery flight phase, we do not change the value of $\theta_{g}$ from its value at $t = t_e$.

\section{RESULTS}

As an example of the wedged object extracting using a UAM, we conduct plug-pulling experiments with a UAM configured with a quadrotor and a 3-DOF robotic arm including a gripper. 
This section compares the results of the aerial plug-pulling experiments with and without the recovery flight control strategy. 

\subsection{Experimental Setups}

\begin{figure}[!t] 
    \centering
    \includegraphics[width=0.45\textwidth]{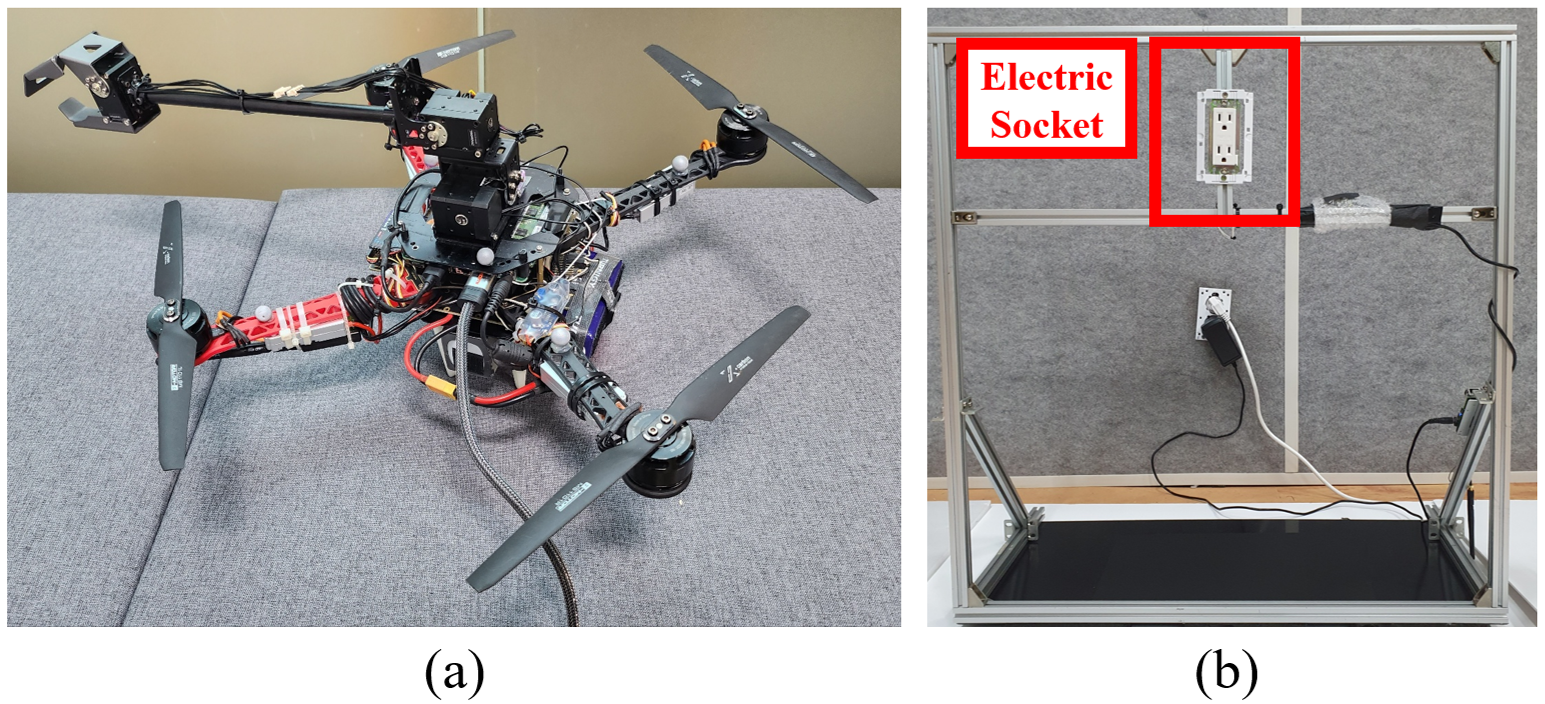}
    \vspace{-0.3cm}
    \caption{(a) UAM utilized for our plug-pulling experiments. (b) Structure of the electric socket.}
    \label{fig: UAM and socket}
\end{figure}

For the UAM, we use the platform configured with an underactuated quadrotor and a 3-DOF robotic arm with a gripper shown in Fig. \ref{fig: UAM and socket}a. 
The quadrotor is made up of the off-the-shelf frame DJI S500, four U3-KV700 motors with corresponding ALPHA-40A-LV electronic speed controllers (ESCs), four 11-inch T-Motor carbon fiber propellers, a 4S Turnigy Lipo battery with Intel NUC for computing, and a 6S Polytronics Lipo battery for the power supplement of the four motors and robotic arm. 
On Intel NUC, the algorithm for the reference trajectory generation of the position and joint angles, a position controller, and a navigation algorithm using the data obtained by OptiTrack are executed with Robot Operating System (ROS) installed in Ubuntu 18.04. 
Based on the total thrust and the desired Euler angles generated by the position controller, the motor commands are transmitted to the four ESCs by Pixhawk 4 which is connected to Intel NUC. 
The robotic arm is configured with an XM540 servo motor, two XM430 servo motors, and an FR12-G101GM gripper manufactured by ROBOTIS dynamixel and is controlled by U2D2. 
For the plug-pulling setting, a stand for a 110 V socket firmly attached to a black plastic plate is installed as in Fig. \ref{fig: UAM and socket}b. 

\subsection{Parameter and Controller Gain Settings}

\subsubsection{Haptic Device}

To control the haptic device, we use the parameters and controller gains shown in Table {\ref{table: parameters and controller gains utilized for the operation of the haptic device}}.

\begin{table}[h]
\caption{Parameters and Controller Gains for the Haptic Device.}
\label{table: parameters and controller gains utilized for the operation of the haptic device}
\begin{center}
\begin{tabular}{|c|c|c|c|}
\hline
$\boldsymbol{M_{H,d}}$ & \multicolumn{3}{c|}{$\boldsymbol{D_{H,d}}$}
\\
\hline
diag$\{0.20, 0.20, 0.20, 0.20\}$ & \multicolumn{3}{c|}{diag$\{0.10, 0.50, 0.10, 0.10\}$} \\
\hline
$\boldsymbol{K_{I,H}}$ & \multicolumn{3}{c|}{$\boldsymbol{K_{fb}}$}
\\
\hline
diag$\{30.0, 30.0, 30.0, 30.0\}$ & \multicolumn{3}{c|}{diag$\{3.00, 1.00, 7.50, 7.50\}$} \\
\hline
$\boldsymbol{K_{recovery}}$ & $k_{d,g}$ & $k_{fb,g}$ & $k_{\tau}$ \\
\hline
diag$\{6.00, 2.00, 15.0, 15.0\}$ & 0.50 & 8.00 & 15.0 \\
\hline
\end{tabular}
\end{center}
\end{table}

\subsubsection{UAM}

For the reference trajectory generation and control of the UAM, we utilize the parameters and controller gains in Table \ref{table: parameters and controller gains utilized for the operation of the UAM}.
\begin{table}[h]
\caption{Parameters and controller gains utilized for the UAM}
\label{table: parameters and controller gains utilized for the operation of the UAM}
\begin{center}
\begin{tabular}{|c|c|c|c|c|}
\hline
$v_{\textrm{max},x}$ & $v_{\textrm{max},y}$ & $v_{\textrm{max},z}$ & $p_{H,\textrm{max},x}$ & $p_{H,\textrm{max},y}$\\
\hline
0.4 & 0.4 & 0.4 & 0.2 & 0.2\\
\hline
$p_{H,\textrm{max},z}$ & $\hat{f}_{A,E,\textrm{th}}$ & $T_e$ & $\hat{m}_t$ & $\hat{g}$ \\
\hline
0.2 & 7.50 & 5.0 & 2.50 & 9.81 \\
\hline
\multicolumn{2}{|c|}{$\boldsymbol{K_P}$} & \multicolumn{2}{|c|}{$\boldsymbol{K_D}$} & $\nu_c$  \\
\hline 
\multicolumn{2}{|c|}{diag$\{4.00, 4.00, 8.00\}$} & \multicolumn{2}{c|}{diag$\{3.00, 3.00, 4.80\}$} & 0.90 \\
\hline
\multicolumn{2}{|c|}{$\boldsymbol{K_R}$} & \multicolumn{3}{c|}{$\boldsymbol{\Gamma_{c,\zeta}}$ = $\boldsymbol{\Gamma_{c,\chi}}$}
\\
\hline
\multicolumn{2}{|c|}{diag$\{12.0, 12.0, 10.0\}$} & \multicolumn{3}{c|}{diag$\{1.00, 1.00, 1.00\}$} \\
\hline
\end{tabular}
\end{center}
\end{table}
However, since the excessively fast estimation of $\boldsymbol{f_{A,E}}$ might induce the undesirable peak in $\boldsymbol{\hat{f}_{A,E}}$ so that the control performance is degraded, we reduce $\nu_c$ from 0.90 to 0.20 only for the estimation of $\boldsymbol{f_{A,E}}$.

\subsection{Scenario}

First, we manipulate our haptic device to drive the UAM toward the 110V plug attached to the electric socket. 
Second, when the gripper of the UAM envelopes the plug-grabbing region, we close the gripper of the UAM by closing that of the haptic device by hand.
Third, we pull the haptic device to make the UAM try to extract the plug from the socket.

To validate our proposed teleoperation strategy, we conduct two experiments: One with the baseline method, and another with the proposed method. 
The baseline method is simply utilizing the teleoperation strategy during the nominal flight for the entire steps of our scenario. 
Otherwise, the proposed method means that we utilize all of the methods introduced in Sections III - V. 

\subsection{Results and Discussion}

\begin{figure}[!t] 
    \centering
    \includegraphics[width=0.50\textwidth]{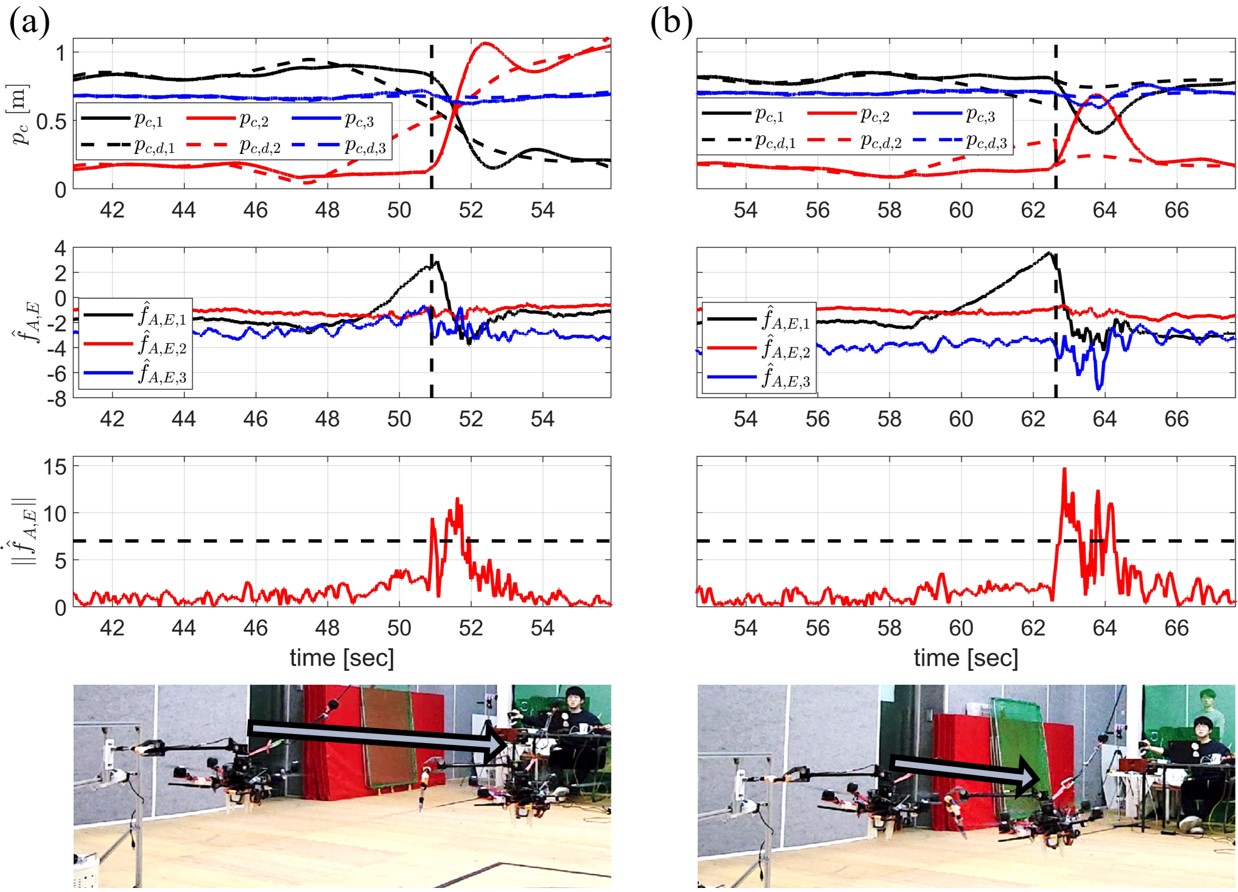}
    \vspace{-0.3cm}
    \caption{History of the UAM's COM position ($\boldsymbol{p_c}$), estimation of the force exerted on the UAM ($\boldsymbol{\hat{f}_{A,E}}$), speed of $\boldsymbol{\hat{f}_{A,E}}$ ($\|\boldsymbol{\dot{\hat{f}}_{A,E}}\|$), and snapshots. The vertical black dashed lines represent the time instant of plug separation. In the plots of $\boldsymbol{p_c}$ and $\boldsymbol{\hat{f}_{A,E}}$, the black, red, and blue lines represent the first, second and third elements, respectively, and the solid and dashed lines mean the actual and desired trajectories, respectively. For the speed of $\boldsymbol{\hat{f}_{A,E}}$, the red line represents the calculated value while the black dashed horizontal lines mean $\dot{\hat{f}}_{A,E,thres}$. (a) Results with the baseline method and (b) results with the \textbf{proposed} method.}
    \label{fig: result plots}
\end{figure}

Fig. \ref{fig: result plots}a presents the results with the baseline teleoperation method. 
The results of $\boldsymbol{\hat{f}_{A,E}}$ and $\|\boldsymbol{\dot{\hat{f}}_{A,E}}\|$ shows an abrupt change in the interaction force after the plug separation. 
Also, from the plot of $\boldsymbol{p_c}$, we can observe that there occur the continuous increases of $\boldsymbol{p_{c,d,1}}$, $\boldsymbol{p_{c,d,2}}$ and $\boldsymbol{p_{c,d,3}}$ along with the pulling direction even after the plug separation due to the limited reaction time of the human operator.
It leads to a large difference between $\boldsymbol{p_c}$ and $\boldsymbol{p_e}$ and prevents the UAM from quickly returning to the plug-pulling location.

Fig. \ref{fig: result plots}b reports the results with the proposed teleoperation method. 
As in Fig. \ref{fig: result plots}a, the results of $\boldsymbol{\hat{f}_{A,E}}$ and $\|\boldsymbol{\dot{\hat{f}}_{A,E}}\|$ also validate that an abrupt change in the pulling force occurs after the plug separation. 
Compared to the history of $\boldsymbol{p_c}$ with the baseline method, we can notice that the proposed method helps the UAM recover the plug-pulling position within a short time. 
Also, even though there also exists an overshoot after the plug separation due to the time delay in estimating $\boldsymbol{\hat{f}_{A,E}}$, from Table \ref{table: quantitative results}, we can note that the magnitude of overshoot in $\boldsymbol{p_c}$ is remarkably reduced compared to the previous result.
From this quantitative result, with the same controller setting for the UAM, we confirm that the proposed teleoperation strategy helps reduce the overshoot in the UAM's position after the plug is extracted from the socket.  

\begin{table}[h]
\caption{Quantitative Results of Overshoot After Plug Separation. Best value is highlighted with bold.}
\label{table: quantitative results}
\begin{center}
\begin{tabular}{|c|c|c|}
\hline
& Baseline & Proposed
\\
\hline
$\underset{t_e \leq t \leq t_e + T_e}{\textrm{max}}\|\boldsymbol{p_c}(t) - \boldsymbol{p_e}\|$ [m] & 0.8934 & \textbf{0.6422} \\
\hline
\end{tabular}
\end{center}
\end{table}

\begin{figure}[!t] 
    \centering
    \includegraphics[width=0.50\textwidth]{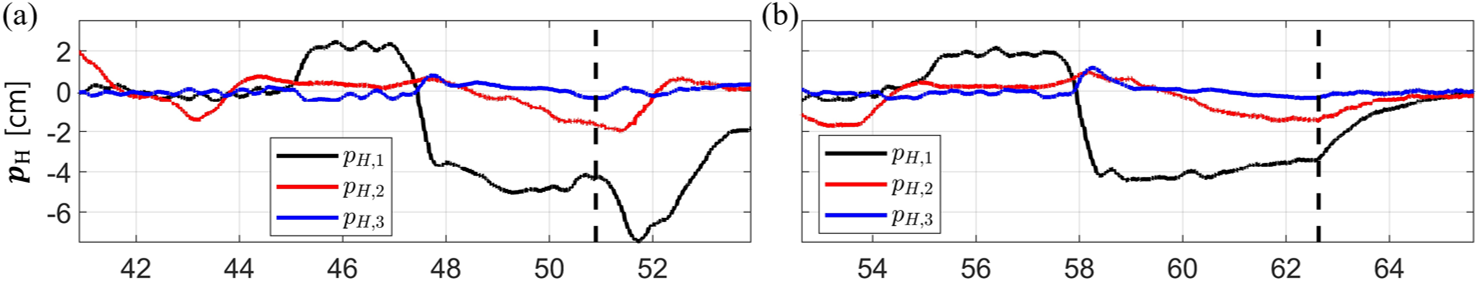}
    \vspace{-0.3cm}
    \caption{History of $\boldsymbol{p_H}$ of the haptic device. The blue, red, and yellow lines represent the trajectory of $p_{H,1}$, $p_{H,2}$ and $p_{H,3}$, respectively. The vertical black dashed line expresses the time instant of the plug separation. (a) Results with the baseline method and (b) results with the \textbf{proposed} method.}
    \label{fig: haptic_result_plot}
\end{figure}

In Fig. \ref{fig: haptic_result_plot}a, it can be seen that $\boldsymbol{p_H}$ slowly returns to $\boldsymbol{0}_{3\times1}$ due to the limitation of the human operator's reaction time. 
This result leads to the excessive overshoot in $\boldsymbol{p_c}$. 
On the other hand, we can notice that $\boldsymbol{p_H}$ automatically returns to its original point in Fig. \ref{fig: haptic_result_plot}b. 
Since $\boldsymbol{p_H}$ becomes $\boldsymbol{0_{3\times1}}$ after the plug separation, the human operator can manipulate the haptic device from its initial configuration when the nominal flight phase is revisited. 

\section{CONCLUSIONS}

In this paper, we propose a haptic-based bilateral teleoperation strategy to compensate for human reaction time for a UAM extracting a wedged object from a static structure (i.e., plug-pulling), which incurs an abrupt decrease in the interaction force.
To this end, the haptic device composed of a 4-DOF robotic arm and a gripper is manufactured for the teleoperation of aerial wedged object-extracting tasks, and their dynamic models are derived. 
Then, we introduce the haptic-based bilateral teleoperation strategy to execute the UAM by the haptic device during the nominal flight phase. 
Also, we construct the algorithm for detecting the extraction of the wedged object, and the reference trajectory generation method for the avoidance of destabilization or an excessive overshoot in the position of the UAM is designed.
As an example of wedged object-extracting tasks, we conduct plug-pulling experiments with a quadrotor-based aerial manipulator. 
From the comparative experiments with the baseline and proposed teleoperation strategies, the results indicate that the proposed one ensures fast recovery to its initial position after the wedged object extraction and reduces the overshoot in the aerial manipulator's position. For future works, we may adopt a hybrid control strategy or model predictive controller for more reduction of the overshoot in the position of the UAM. 
Also, an additional device for visual or tactile feedback may be attached to our UAM.


\addtolength{\textheight}{-12cm}   





\bibliographystyle{IEEEtran}
\bibliography{root}

\begin{thebibliography}{10}
\providecommand{\url}[1]{#1}
\csname url@samestyle\endcsname
\providecommand{\newblock}{\relax}
\providecommand{\bibinfo}[2]{#2}
\providecommand{\BIBentrySTDinterwordspacing}{\spaceskip=0pt\relax}
\providecommand{\BIBentryALTinterwordstretchfactor}{4}
\providecommand{\BIBentryALTinterwordspacing}{\spaceskip=\fontdimen2\font plus
\BIBentryALTinterwordstretchfactor\fontdimen3\font minus \fontdimen4\font\relax}
\providecommand{\BIBforeignlanguage}[2]{{%
\expandafter\ifx\csname l@#1\endcsname\relax
\typeout{** WARNING: IEEEtran.bst: No hyphenation pattern has been}%
\typeout{** loaded for the language `#1'. Using the pattern for}%
\typeout{** the default language instead.}%
\else
\language=\csname l@#1\endcsname
\fi
#2}}
\providecommand{\BIBdecl}{\relax}
\BIBdecl

\bibitem{kim2015operating}
S.~Kim, H.~Seo, and H.~J. Kim, ``Operating an unknown drawer using an aerial manipulator,'' in \emph{2015 IEEE International Conference on Robotics and Automation (ICRA)}.\hskip 1em plus 0.5em minus 0.4em\relax IEEE, 2015, pp. 5503--5508.

\bibitem{car2018impedance}
M.~Car, A.~Ivanovic, M.~Orsag, and S.~Bogdan, ``Impedance based force control for aerial robot peg-in-hole insertion tasks,'' in \emph{2018 IEEE/RSJ International Conference on Intelligent Robots and Systems (IROS)}.\hskip 1em plus 0.5em minus 0.4em\relax IEEE, 2018, pp. 6734--6739.

\bibitem{trujillo2019novel}
M.~{\'A}. Trujillo, J.~R. Mart{\'\i}nez-de Dios, C.~Mart{\'\i}n, A.~Viguria, and A.~Ollero, ``Novel aerial manipulator for accurate and robust industrial ndt contact inspection: A new tool for the oil and gas inspection industry,'' \emph{Sensors}, vol.~19, no.~6, p. 1305, 2019.

\bibitem{lee2020aerial}
D.~Lee, H.~Seo, D.~Kim, and H.~J. Kim, ``Aerial manipulation using model predictive control for opening a hinged door,'' in \emph{2020 IEEE International Conference on Robotics and Automation (ICRA)}.\hskip 1em plus 0.5em minus 0.4em\relax IEEE, 2020, pp. 1237--1242.

\bibitem{sun2021switchable}
Y.~Sun, Z.~Jing, P.~Dong, J.~Huang, W.~Chen, and H.~Leung, ``A switchable unmanned aerial manipulator system for window-cleaning robot installation,'' \emph{IEEE Robotics and Automation Letters}, vol.~6, no.~2, pp. 3483--3490, 2021.

\bibitem{kim2017robust}
S.~Kim, S.~Choi, H.~Kim, J.~Shin, H.~Shim, and H.~J. Kim, ``Robust control of an equipment-added multirotor using disturbance observer,'' \emph{IEEE Transactions on Control Systems Technology}, vol.~26, no.~4, pp. 1524--1531, 2017.

\bibitem{liang2021low}
J.~Liang, Y.~Chen, N.~Lai, B.~He, Z.~Miao, and Y.~Wang, ``Low-complexity prescribed performance control for unmanned aerial manipulator robot system under model uncertainty and unknown disturbances,'' \emph{IEEE Transactions on Industrial Informatics}, vol.~18, no.~7, pp. 4632--4641, 2021.

\bibitem{lee2022rise}
D.~Lee, J.~Byun, and H.~J. Kim, ``Rise-based trajectory tracking control of an aerial manipulator under uncertainty,'' \emph{IEEE Control Systems Letters}, vol.~6, pp. 3379--3384, 2022.

\bibitem{mersha2013bilateral}
A.~Y. Mersha, S.~Stramigioli, and R.~Carloni, ``On bilateral teleoperation of aerial robots,'' \emph{IEEE Transactions on Robotics}, vol.~30, no.~1, pp. 258--274, 2013.

\bibitem{byun2023hybrid}
J.~Byun, I.~Jang, D.~Lee, and H.~J. Kim, ``A hybrid controller enhancing transient performance for an aerial manipulator extracting a wedged object,'' \emph{IEEE Transactions on Automation Science and Engineering}, 2023.

\bibitem{stramigioli2010novel}
S.~Stramigioli, R.~Mahony, and P.~Corke, ``A novel approach to haptic tele-operation of aerial robot vehicles,'' in \emph{2010 IEEE International Conference on Robotics and Automation}.\hskip 1em plus 0.5em minus 0.4em\relax IEEE, 2010, pp. 5302--5308.

\bibitem{macchini2020hand}
M.~Macchini, T.~Havy, A.~Weber, F.~Schiano, and D.~Floreano, ``Hand-worn haptic interface for drone teleoperation,'' in \emph{2020 IEEE International Conference on Robotics and Automation (ICRA)}.\hskip 1em plus 0.5em minus 0.4em\relax IEEE, 2020, pp. 10\,212--10\,218.

\bibitem{masone2018shared}
C.~Masone, M.~Mohammadi, P.~Robuffo~Giordano, and A.~Franchi, ``Shared planning and control for mobile robots with integral haptic feedback,'' \emph{The International Journal of Robotics Research}, vol.~37, no.~11, pp. 1395--1420, 2018.

\bibitem{coelho2020whole}
A.~Coelho, H.~Singh, K.~Kondak, and C.~Ott, ``Whole-body bilateral teleoperation of a redundant aerial manipulator,'' in \emph{2020 IEEE International Conference on Robotics and Automation (ICRA)}.\hskip 1em plus 0.5em minus 0.4em\relax IEEE, 2020, pp. 9150--9156.

\bibitem{gioioso2015force}
G.~Gioioso, M.~Mohammadi, A.~Franchi, and D.~Prattichizzo, ``A force-based bilateral teleoperation framework for aerial robots in contact with the environment,'' in \emph{2015 IEEE International Conference on Robotics and Automation (ICRA)}.\hskip 1em plus 0.5em minus 0.4em\relax IEEE, 2015, pp. 318--324.

\bibitem{lee2020visual}
J.~Lee, R.~Balachandran, Y.~S. Sarkisov, M.~De~Stefano, A.~Coelho, K.~Shinde, M.~J. Kim, R.~Triebel, and K.~Kondak, ``Visual-inertial telepresence for aerial manipulation,'' in \emph{2020 IEEE International Conference on Robotics and Automation (ICRA)}.\hskip 1em plus 0.5em minus 0.4em\relax IEEE, 2020, pp. 1222--1229.

\bibitem{kim2020human}
D.~Kim and P.~Y. Oh, ``Human-drone interaction for aerially manipulated drilling using haptic feedback,'' in \emph{2020 IEEE/RSJ International Conference on Intelligent Robots and Systems (IROS)}.\hskip 1em plus 0.5em minus 0.4em\relax IEEE, 2020, pp. 9774--9780.

\bibitem{allenspach2022towards}
M.~Allenspach, N.~Lawrance, M.~Tognon, and R.~Siegwart, ``Towards 6dof bilateral teleoperation of an omnidirectional aerial vehicle for aerial physical interaction,'' in \emph{2022 International Conference on Robotics and Automation (ICRA)}.\hskip 1em plus 0.5em minus 0.4em\relax IEEE, 2022, pp. 9302--9308.

\bibitem{yang2014dynamics}
H.~Yang and D.~Lee, ``Dynamics and control of quadrotor with robotic manipulator,'' in \emph{2014 IEEE international conference on robotics and automation (ICRA)}.\hskip 1em plus 0.5em minus 0.4em\relax IEEE, 2014, pp. 5544--5549.

\bibitem{ha2018disturbance}
W.~Ha and J.~Back, ``A disturbance observer-based robust tracking controller for uncertain robot manipulators,'' \emph{International Journal of Control, Automation and Systems}, vol.~16, pp. 417--425, 2018.

\bibitem{tomic2017external}
T.~Tomi{\'c}, C.~Ott, and S.~Haddadin, ``External wrench estimation, collision detection, and reflex reaction for flying robots,'' \emph{IEEE Transactions on Robotics}, vol.~33, no.~6, pp. 1467--1482, 2017.

\bibitem{liu2021toward}
Z.~Liu and K.~Karydis, ``Toward impact-resilient quadrotor design, collision characterization and recovery control to sustain flight after collisions,'' in \emph{2021 IEEE International Conference on Robotics and Automation (ICRA)}.\hskip 1em plus 0.5em minus 0.4em\relax IEEE, 2021, pp. 183--189.

\end{thebibliography}

\end{document}